\documentclass[letterpaper]{article} 
\usepackage{aaai23}  
\usepackage{times}  
\usepackage{helvet}  
\usepackage{courier}  
\usepackage[hyphens]{url}  
\usepackage{graphicx} 
\usepackage{natbib}  
\usepackage{caption} 
\usepackage{algorithm}
\usepackage{algorithmic}

\usepackage{newfloat}
\usepackage{listings}
\DeclareCaptionStyle{ruled}{labelfont=normalfont,labelsep=colon,strut=off} 
\floatstyle{ruled}
\newfloat{listing}{tb}{lst}{}
\floatname{listing}{Listing}
\nocopyright
\usepackage{amsmath}

\usepackage{booktabs}
\usepackage{multirow}

\title{TrOCR: Transformer-based Optical Character Recognition\\ with Pre-trained Models}

\author{Minghao Li$^{1}$\thanks{Work done during internship at Microsoft Research Asia.}, Tengchao Lv$^{2}$, Jingye Chen$^{2}$$^{*}$, Lei Cui$^{2}$, \\
Yijuan Lu$^{2}$, Dinei Florencio$^{2}$, Cha Zhang$^{2}$, Zhoujun Li$^{1}$, Furu Wei$^{2}$}

\affiliations{
$^{1}$Beihang University\\
$^{2}$Microsoft Corporation\\
\{liminghao1630, lizj\}@buaa.edu.cn\\
\{tengchaolv, v-jingyechen, lecu, yijlu, dinei, chazhang, fuwei\}@microsoft.com\\
}

\begin{document}

\maketitle

\begin{abstract}
Text recognition is a long-standing research problem for document digitalization. Existing approaches are usually built based on CNN for image understanding and RNN for char-level text generation. In addition, another language model is usually needed to improve the overall accuracy as a post-processing step. In this paper, we propose an end-to-end text recognition approach with pre-trained image Transformer and text Transformer models, namely \textbf{TrOCR}, which leverages the Transformer architecture for both image understanding and wordpiece-level text generation. The TrOCR model is simple but effective, and can be pre-trained with large-scale synthetic data and fine-tuned with human-labeled datasets. Experiments show that the TrOCR model outperforms the current state-of-the-art models on the printed, handwritten and scene text recognition tasks. The TrOCR models and code are publicly available at \url{https://aka.ms/trocr}.
\end{abstract}

\section{Introduction}

Optical Character Recognition (OCR) is the electronic or mechanical conversion of images of typed, handwritten or printed text into machine-encoded text, whether from a scanned document, a photo of a document, a scene photo or from subtitle text superimposed on an image. Typically, an OCR system includes two main modules: a text detection module and a text recognition module. Text detection aims to localize all text blocks within the text image, either at word-level or textline-level. The text detection task is usually considered as an object detection problem where conventional object detection models such as YoLOv5 and DBNet~\cite{liao2019realtime} can be applied. Meanwhile, text recognition aims to understand the text image content and transcribe the visual signals into natural language tokens. The text recognition task is usually framed as an encoder-decoder problem where existing methods leveraged CNN-based encoder for image understanding and RNN-based decoder for text generation. In this paper, we focus on the text recognition task for document images and leave text detection as the future work.

Recent progress in text recognition~\citep{diaz2021rethinking} has witnessed the significant improvements by taking advantage of the Transformer~\cite{vaswani2017attention} architectures. However, existing methods are still based on CNNs as the backbone, where the self-attention is built on top of CNN backbones as encoders to understand the text image. For decoders, Connectionist Temporal Classification (CTC) \cite{graves2006connectionist} is usually used compounded with an external language model on the character-level to improve the overall accuracy. Despite the great success achieved by the hybrid encoder/decoder method, there is still a lot of room to improve with pre-trained CV and NLP models: 1) the network parameters in existing methods are trained from scratch with synthetic/human-labeled datasets, leaving large-scale pre-trained models unexplored. 2) as image Transformers become more and more popular~\citep{dosovitskiy2020vit,touvron2020deit}, especially the recent self-supervised image pre-training~\citep{bao2021beit}, it is straightforward to investigate whether pre-trained image Transformers can replace CNN backbones, meanwhile exploiting the pre-trained image Transformers to work together with the pre-trained text Transformers in a single framework on the text recognition task. 

\begin{figure*}[t]
\centering
\includegraphics[width=0.85\textwidth]{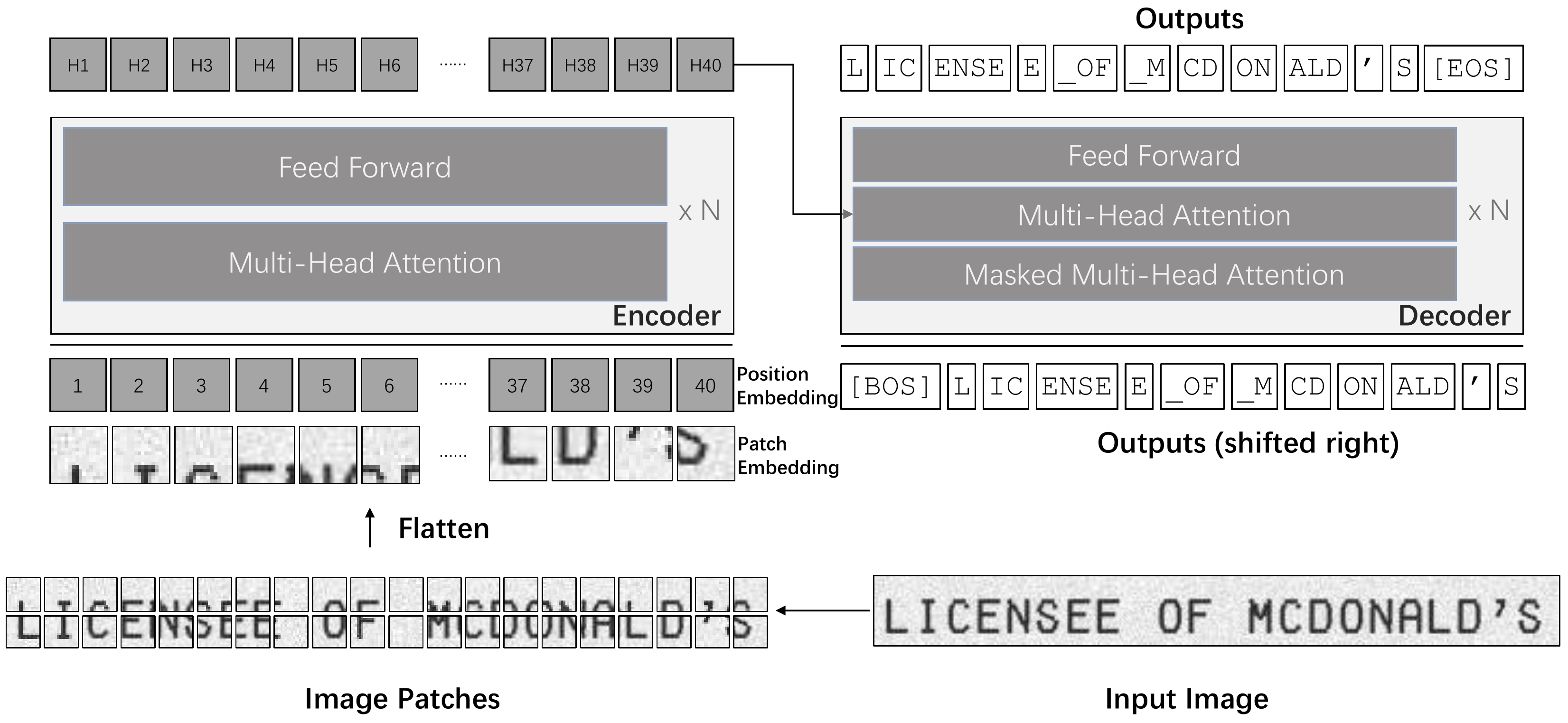}
\caption{The architecture of TrOCR, where an encoder-decoder model is designed with a pre-trained image Transformer as the encoder and a pre-trained text Transformer as the decoder.}
\label{fig:model}
\end{figure*}

To this end, we propose \textbf{TrOCR}, an end-to-end Transformer-based OCR model for text recognition with pre-trained CV and NLP models, which is shown in Figure \ref{fig:model}. Distinct from the existing text recognition models, TrOCR is a simple but effective model which does not use the CNN as the backbone. Instead, following~\citep{dosovitskiy2020vit}, it first resizes the input text image into $384\times384$ and then the image is split into a sequence of $16\times16$ patches which are used as the input to image Transformers. Standard Transformer architecture with the self-attention mechanism is leveraged on both encoder and decoder parts, where wordpiece units are generated as the recognized text from the input image. To effectively train the TrOCR model, the encoder can be initialized with pre-trained ViT-style models~\citep{dosovitskiy2020vit,touvron2020deit,bao2021beit} while the decoder can be initialized with pre-trained BERT-style models~\citep{devlin2019bert,liu2019roberta,dong2019unified, wang2020minilm}, respectively. Therefore, the advantage of TrOCR is three-fold. First, TrOCR uses the pre-trained image Transformer and text Transformer models, which take advantages of large-scale unlabeled data for image understanding and language modeling, with no need for an external language model. Second, TrOCR does not require any convolutional network for the backbone and does not introduce any image-specific inductive biases, which makes the model very easy to implement and maintain. Finally, experiment results on OCR benchmark datasets show that the TrOCR can achieve state-of-the-art results on printed, handwritten and scene text image datasets without any complex pre/post-processing steps. Furthermore, we can easily extend the TrOCR for multilingual text recognition with minimum efforts, where just leveraging multilingual pre-trained models in the decoder-side and expand the dictionary.

The contributions of this paper are summarized as follows:
\begin{enumerate}
    \item We propose TrOCR, an end-to-end Transformer-based OCR model for text recognition with pre-trained CV and NLP models. To the best of our knowledge, this is the first work that jointly leverages pre-trained image and text Transformers for the text recognition task in OCR. 
    \item TrOCR achieves state-of-the-art results with a standard Transformer-based encoder-decoder model, which is convolution free and does not rely on any complex pre/post-processing steps.
    \item The TrOCR models and code are publicly available at \url{https://aka.ms/trocr}.
\end{enumerate}

\section{TrOCR}

\subsection{Model Architecture}
TrOCR is built up with the Transformer architecture, including an image Transformer for extracting the visual features and a text Transformer for language modeling. We adopt the vanilla Transformer encoder-decoder structure in TrOCR. The encoder is designed to obtain the representation of the image patches and the decoder is to generate the wordpiece sequence with the guidance of the visual features and previous predictions. 

\subsubsection{Encoder}
The encoder receives an input image $x_{\rm img} \in\Re^{3\times H_0\times W_0}$, and resizes it to a fixed size $(H, W)$. Since the Transformer encoder cannot process the raw images unless they are a sequence of input tokens, the encoder decomposes the input image into a batch of $N=HW/P^2$ foursquare patches with a fixed size of $(P, P)$,
while the width $W$ and the height $H$ of the resized image are guaranteed to be divisible by the patch size $P$. Subsequently, the patches are flattened into vectors and linearly projected to $D$-dimensional vectors, aka the patch embeddings. $D$ is the hidden size of the Transformer through all of its layers.

Similar to ViT \cite{dosovitskiy2020vit} and DeiT \cite{touvron2020deit}, we keep the special token ``[CLS]'' that is usually used for image classification tasks. The ``[CLS]'' token brings together all the information from all the patch embeddings and represents the whole image. Meanwhile, we also keep the distillation token in the input sequence when using the DeiT pre-trained models for encoder initialization, which allows the model to learn from the teacher model. The patch embeddings and two special tokens are given learnable 1D position embeddings according to their absolute positions. 

Unlike the features extracted by the CNN-like network, the Transformer models have no image-specific inductive biases and process the image as a sequence of patches, which makes the model easier to pay different attention to either the whole image or the independent patches.

\subsubsection{Decoder}
\label{sec:decoder}

We use the original Transformer decoder for TrOCR. The standard Transformer decoder also has a stack of identical layers, which have similar structures to the layers in the encoder, except that the decoder inserts the ``encoder-decoder attention'' between the multi-head self-attention and feed-forward network to distribute different attention on the output of the encoder. In the encoder-decoder attention module, the keys and values come from the encoder output, while the queries come from the decoder input. In addition, the decoder leverages the attention masking in the self-attention to prevent itself from getting more information during training than prediction. Based on the fact that the output of the decoder will right shift one place from the input of the decoder, the attention mask needs to ensure the output for the position $i$ can only pay attention to the previous output, which is the input on the positions less than $i$:

\begin{align}
    h_i &= Proj(Emb(Token_i))\notag \\
    \sigma(h_{ij}) &= \frac{e^{h_{ij}}}{\sum_{k=1}^V e^{h_{ik}}} \ \ \ for\ j=1,2,\dots,V \notag
\end{align}

The hidden states from the decoder are projected by a linear layer from the model dimension to the dimension of the vocabulary size $V$, while the probabilities over the vocabulary are calculated on that by the softmax function. We use beam search to get the final output.

\subsection{Model Initialization}
Both the encoder and the decoder are initialized by the public models pre-trained on large-scale labeled and unlabeled datasets.

\subsubsection{Encoder Initialization}

The DeiT~\cite{touvron2020deit} and BEiT~\cite{bao2021beit} models are used for the encoder initialization in the TrOCR models.
DeiT trains the image Transformer with ImageNet \cite{deng2009imagenet} as the sole training set. The authors try different hyper-parameters and data augmentation to make the model data-efficient. Moreover, they distill the knowledge of a strong image classifier to a distilled token in the initial embedding, which leads to a competitive result compared to the CNN-based models.

Referring to the Masked Language Model pre-training task, BEiT proposes the Masked Image Modeling task to pre-train the image Transformer. Each image will be converted to two views: image patches and visual tokens. They tokenize the original image into visual tokens by the latent codes of discrete VAE \cite{ramesh2021zero}, randomly mask some image patches, and make the model recover the original visual tokens. The structure of BEiT is the same as the image Transformer and lacks the distilled token when compared with DeiT.



\subsubsection{Decoder Initialization}

We use the RoBERTa \cite{liu2019roberta} models and the MiniLM \cite{wang2020minilm} models to initialize the decoder. Generally, RoBERTa is a replication study of \cite{devlin2019bert} that carefully measures the impact of many key hyperparameters and training data size. Based on BERT, they remove the next sentence prediction objective and dynamically change the masking pattern of the Masked Language Model. 

The MiniLM are compressed models of the large pre-trained Transformer models while retaining 99\% performance. Instead of using the soft target probabilities of masked language modeling predictions or intermediate representations of the teacher models to guide the training of the student models in the previous work. The MiniLM models are trained by distilling the self-attention module of the last Transformer layer of the teacher models and introducing a teacher assistant to assist with the distillation. 

When loading the above models to the decoders, the structures do not precisely match since both of them are only the encoder of the Transformer architecture. For example, the encoder-decoder attention layers are absent in these models.
To address this, we initialize the decoders with the RoBERTa and MiniLM models by manually setting the corresponding parameter mapping, and the absent parameters are randomly initialized.

\subsection{Task Pipeline}
In this work, the pipeline of the text recognition task is that given the textline images, the model extracts the visual features and predicts the wordpiece tokens relying on the image and the context generated before.
The sequence of ground truth tokens is followed by an ``[EOS]'' token, which indicates the end of a sentence. During training, we shift the sequence backward by one place and add the ``[BOS]'' token to the beginning indicating the start of generation. The shifted ground truth sequence is fed into the decoder, and the output of that is supervised by the original ground truth sequence with the cross-entropy loss. For inference, the decoder starts from the ``[BOS]'' token to predict the output iteratively while continuously taking the newly generated output as the next input.

\subsection{Pre-training}
We use the text recognition task for the pre-training phase, since this task can make the models learn the knowledge of both the visual feature extraction and the language model. The pre-training process is divided into two stages that differ by the used dataset. In the first stage, we synthesize a large-scale dataset consisting of hundreds of millions of printed textline images and pre-train the TrOCR models on that. 
In the second stage, we build two relatively small datasets corresponding to printed and handwritten downstream tasks, containing millions of textline images each. We use the existed and widely adopted synthetic scene text datasets for the scene text recognition task. 
Subsequently, we pre-train separate models on these task-specific datasets in the second stage, all initialized by the first-stage model. 

\subsection{Fine-tuning}
Except for the experiments regarding scene text recognition, the pre-trained TrOCR models are fine-tuned on the downstream text recognition tasks. The outputs of the TrOCR models are based on Byte Pair Encoding (BPE)~\cite{sennrich2015neural} and SentencePiece~\cite{kudo2018sentencepiece} and do not rely on any task-related vocabularies. 


\begin{table}[ht]
\centering
\small
\begin{tabular}{ccccc}
\hline
\textbf{Encoder} & \textbf{Decoder}  & \textbf{Precision} & \textbf{Recall} & \textbf{F1}    \\
\hline
$\textrm{DeiT}_{\rm BASE}$             & $\textrm{RoBERTa}_{\rm BASE}$  & 69.28              & 69.06           & 69.17          \\
$\textrm{BEiT}_{\rm BASE}$             & $\textrm{RoBERTa}_{\rm BASE}$  & 76.45              & 76.18           & 76.31          \\
ResNet50         & $\textrm{RoBERTa}_{\rm BASE}$  & 66.74              & 67.29           & 67.02          \\
$\textrm{DeiT}_{\rm BASE}$             & $\textrm{RoBERTa}_{\rm LARGE}$ & 77.03              & 76.53           & 76.78          \\
$\textrm{BEiT}_{\rm BASE}$             & $\textrm{RoBERTa}_{\rm LARGE}$ & \textbf{79.67}     & \textbf{79.06}  & \textbf{79.36} \\
ResNet50         & $\textrm{RoBERTa}_{\rm LARGE}$ & 72.54              & 71.13           & 71.83          \\
\hline
\end{tabular}
\caption{Ablation study on the SROIE dataset, where all the models are trained using the SROIE dataset only.}
\label{tab:ablation}
\end{table}

\begin{table}
\centering
\begin{tabular}{lccc}
\hline
\textbf{Model}          & \textbf{Precision} & \textbf{Recall} & \textbf{F1}    \\ \hline
From Scratch            & 38.06              & 38.43           & 38.24          \\
+ Pretrained Model      & 72.95              & 72.56           & 72.75          \\
+ Data Augmentation     & 82.58              & 82.03           & 82.30          \\
+ First-Stage Pretrain  & 95.31              & 95.65           & 95.48          \\
+ Second-Stage Pretrain & \textbf{95.76}     & \textbf{95.91}  & \textbf{95.84} \\ \hline
\end{tabular}
\caption{Ablation study of pretrained model initialization, data augmentation and two stages of pre-training on the SROIE dataset.}
\label{tab:ablation2}
\end{table}

\subsection{Data Augmentation}
We leverage data augmentation to enhance the variety of the pre-training and fine-tuning data. Six kinds of image transformations plus keeping the original are taken for printed and handwritten datasets, which are random rotation (-10 to 10 degrees), Gaussian blurring, image dilation, image erosion, downscaling, and underlining. We randomly decide which image transformation to take with equal possibilities for each sample. For scene text datasets, RandAugment~\cite{cubuk2020randaugment} is applied following \cite{atienza2021vision}, and the augmentation types include inversion, curving, blur, noise, distortion, rotation, etc.

\section{Experiments}

\subsection{Data}

\subsubsection{Pre-training Dataset}

To build a large-scale high-quality dataset, we sample two million document pages from the publicly available PDF files on the Internet. Since the PDF files are digital-born, we can get pretty printed textline images by converting them into page images and extracting the textlines with their cropped images.
In total, the first-stage pre-training dataset contains 684M textlines.


We use 5,427 handwritten fonts\footnote{\tiny The fonts are obtained from \url{https://fonts.google.com/?category=Handwriting} and \url{https://www.1001fonts.com/handwritten-fonts.html}.} to synthesize handwritten textline images by the TRDG\footnote{\tiny\url{https://github.com/Belval/TextRecognitionDataGenerator}}, an open-source text recognition data generator. The text used for generation is crawled from random pages of Wikipedia. The handwritten dataset for the second-stage pre-training consists of 17.9M textlines, including IIIT-HWS dataset~\cite{krishnan2016generating}.
In addition, we collect around 53K receipt images in the real world and recognize the text on them by commercial OCR engines. According to the results, we crop the textlines by their coordinates and rectify them into normalized images.
We also use TRDG to synthesize 1M printed textline images with two receipt fonts and the built-in printed fonts. In total, the printed dataset consists of 3.3M textlines.
The second-stage pre-training data for the scene text recognition are MJSynth (MJ)~\cite{synth90ka} and SynthText (ST)~\cite{gupta2016synthetic}, totaling about 16M text images.

\subsubsection{Benchmarks}
The SROIE (Scanned Receipts OCR and Information Extraction) dataset (Task 2) focuses on text recognition in receipt images.
There are 626 receipt images and 361 receipt images in the training and test sets of SROIE.
Since the text detection task is not included in this work, we use cropped images of the textlines for evaluation, which are obtained by cropping the whole receipt images according to the ground truth bounding boxes.

The IAM Handwriting Database is composed of handwritten English text, which is the most popular dataset for handwritten text recognition. We use the Aachen's partition of the dataset\footnote{\tiny\url{https://github.com/jpuigcerver/Laia/tree/master/egs/iam}}: 6,161 lines from 747 forms in the train set, 966 lines from 115 forms in the validation set and 2,915 lines from 336 forms in the test set.

Recognizing scene text images is more challenging than printed text images, as many images in the wild suffer from blur, occlusion, or low-resolution problems. Here we leverage some widely-used benchmarks, including IIIT5K-3000 \cite{mishra2012top}, SVT-647 \cite{wang2011end}, IC13-857, IC13-1015 \cite{karatzas2013icdar}, IC15-1811, IC15-2077 \cite{karatzas2015icdar}, SVTP-645 \cite{phan2013recognizing}, and CT80-288 \cite{risnumawan2014robust} to evaluate the capacity of the proposed TrOCR.

\begin{table}[ht]
\centering
\begin{tabular}{cccc}
\hline
\textbf{Model} & \textbf{Recall} & \textbf{Precision} & \textbf{F1} \\ \hline
CRNN            & 28.71              & 48.58          & 36.09         \\
Tesseract OCR   & 57.50              & 51.93           & 54.57       \\ \hline
H\&H Lab & 96.35           & 96.52              & 96.43       \\
MSOLab   & 94.77           & 94.88              & 94.82       \\
CLOVA OCR & 94.3            & 94.88              & 94.59       \\ \hline

$\textrm{TrOCR}_{\rm SMALL}$     & 95.89           & 95.74              & 95.82       \\
$\textrm{TrOCR}_{\rm BASE}$      & 96.37           & 96.31              & 96.34       \\
$\textrm{TrOCR}_{\rm LARGE}$     & 96.59           & 96.57              & \textbf{96.58}       \\
\hline
\end{tabular}
\caption{Evaluation results (word-level Precision, Recall, F1) on the SROIE dataset, where the baselines come from the SROIE leaderboard (\url{https://rrc.cvc.uab.es/?ch=13&com=evaluation&task=2}).}
\label{tab:sroie}
\end{table}

\subsection{Settings}

The TrOCR models are built upon the Fairseq~\cite{ott2019fairseq} which is a popular sequence modeling toolkit. For the model initialization, the DeiT models are implemented and initialized by the code and the pre-trained models from the timm library \cite{rw2019timm} while the BEiT models and the MiniLM models are from the UniLM’s official repository\footnote{\tiny\url{https://github.com/microsoft/unilm}}. The RoBERTa models come from the corresponding page in the Fairseq GitHub repository.
We use 32 V100 GPUs with the memory of 32GBs for pre-training and 8 V100 GPUs for fine-tuning. For all the models, the batch size is set to 2,048 and the learning rate is 5e-5. We use the BPE and sentencepiece tokenizer from Fairseq to tokenize the textlines to wordpieces.

We employ the $384 \times 384$ resolution and $16 \times 16$ patch size for DeiT and BEiT encoders. The $\textrm{DeiT}_{\rm SMALL}$ has 12 layers with 384 hidden sizes and 6 heads. Both the $\textrm{DeiT}_{\rm BASE}$ and the $\textrm{BEiT}_{\rm BASE}$ have 12 layers with 768 hidden sizes and 12 heads while the $\textrm{BEiT}_{\rm LARGE}$ has 24 layers with 1024 hidden sizes and 16 heads.
We use 6 layers, 256 hidden sizes and 8 attention heads for the small decoders, 512 hidden sizes for the base decoders and 12 layers, 1,024 hidden sizes and 16 heads for the large decoders. For this task, we only use the last half of all layers from the corresponding RoBERTa model, which are the last 6 layers for the $\textrm{RoBERTa}_{\rm BASE}$ and the last 12 layers for the $\textrm{RoBERTa}_{\rm LARGE}$. The beam size is set to 10 for TrOCR models.

We take the CRNN model \cite{shi2016end} as the baseline model. The CRNN model is composed of convolutional layers for image feature extraction, recurrent layers for sequence modeling and the final frame label prediction, and a transcription layer to translate the frame predictions to the final label sequence. To address the character alignment issue, they use the CTC loss to train the CRNN model. 
For a long time, the CRNN model is the dominant paradigm for text recognition.
We use the PyTorch implementation\footnote{\tiny\url{https://github.com/meijieru/crnn.pytorch}} and initialized the parameters by the provided pre-trained model. 

\subsection{Evaluation Metrics}
The SROIE dataset is evaluated using the word-level precision, recall and f1 score. If repeated words appear in the ground truth, they are also supposed to appear in the prediction. The precision, recall and f1 score are described as:

\begin{gather}
Precision=\frac{\text{Correct matches}}{\text{The number of the detected words}}\notag \\
Recall=\frac{\text{Correct matches}}{\text{The number of the ground truth words}}\notag \\
F1=\frac{\text{2} \times \text{Precision} \times \text{Recall}}{\text{Precision + Recall}}\notag.
\end{gather}

The IAM dataset is evaluated by the case-sensitive Character Error Rate (CER).
The scene text datasets are evaluated by the Word Accuracy. For fair comparison, we filter the final output string to suit the popular 36-character charset (lowercase alphanumeric) in this task.


\subsection{Results}

\subsubsection{Architecture Comparison}

We compare different combinations of the encoder and decoder to find the best settings. For encoders, we compare DeiT, BEiT and the ResNet-50 network. Both the DeiT and BEiT are the base models in their original papers. For decoders, we compare the base decoders initialized by $\textrm{RoBERTa}_{\rm BASE}$ and the large decoders initialized by $\textrm{RoBERTa}_{\rm LARGE}$. 
For further comparison, we also evaluate the CRNN baseline model and the Tesseract OCR in this section, while the latter is an open-source OCR Engine using the LSTM network.

Table~\ref{tab:ablation} shows the results of combined models. From the results, we observe that the BEiT encoders show the best performance among the three types of encoders while the best decoders are the $\textrm{RoBERTa}_{\rm LARGE}$ decoders. Apparently, the pre-trained models on the vision task improve the performance of text recognition models, and the pure Transformer models are better than the CRNN models and the Tesseract on this task. According to the results, we mainly use three settings on the subsequent experiments: $\textbf{TrOCR}_{\bf SMALL}$ (total parameters=62M) consists of the encoder of $\textrm{DeiT}_{\rm SMALL}$ and the decoder of MiniLM, $\textbf{TrOCR}_{\bf BASE}$ (total parameters=334M) consists of the encoder of $\textrm{BEiT}_{\rm BASE}$ and the decoder of $\textrm{RoBERTa}_{\rm LARGE}$, $\textbf{TrOCR}_{\bf LARGE}$ (total parameters=558M) consists of the encoder of $\textrm{BEiT}_{\rm LARGE}$ and the decoder of $\textrm{RoBERTa}_{\rm LARGE}$. In Table~\ref{tab:ablation2}, we have also done some ablation experiments to verify the effect of pre-trained model initialization, data augmentation, and two stages of pre-training. All of them have great improvements to the TrOCR models.

\begin{table*}[ht]
\centering
\begin{tabular}{ccccc}
\hline
\textbf{Model}      &\textbf{Architecture}             & \textbf{Training Data}   & \textbf{External LM} & \textbf{CER} \\ \hline

\cite{bluche2017gated}   &   GCRNN / CTC     & Synthetic + IAM          & Yes                  & 3.2          \\
\cite{michael2019evaluating}& LSTM/LSTM w/Attn  & IAM                      & No                   & 4.87         \\ 
\cite{wang2020decoupled}  &    FCN / GRU    & IAM                      & No                   & 6.4          \\

\cite{kang2020pay}      &    Transformer w/ CNN     & Synthetic + IAM          & No                   & 4.67         \\

\cite{diaz2021rethinking} & S-Attn / CTC      & Internal + IAM           & No                  & 3.53      \\
\cite{diaz2021rethinking}  &  S-Attn / CTC    & Internal + IAM          & Yes                  & 2.75         \\
\cite{diaz2021rethinking}  &  Transformer w/ CNN    & Internal + IAM          & No                  & 2.96         \\
\hline
$\textrm{TrOCR}_{\rm SMALL}$            &  Transformer        & Synthetic + IAM          & No                   & 4.22             \\
$\textrm{TrOCR}_{\rm BASE}$             &  Transformer        & Synthetic + IAM          & No                   & 3.42             \\
$\textrm{TrOCR}_{\rm LARGE}$             &   Transformer      & Synthetic + IAM          & No                   & 2.89             \\
\hline
\end{tabular}
\caption{Evaluation results (CER) on the IAM Handwriting dataset.}
\label{tab:iam}
\end{table*}

\begin{table*}
\centering
\begin{tabular}{ccccccc}\hline
\textbf{Model}       & \textbf{Parameters} & \textbf{Total Sentences} & \textbf{Total Tokens} & \textbf{Time}   & \textbf{Speed \#Sentences} & \textbf{Speed \#Tokens} \\
\hline
$\textrm{TrOCR}_{\rm SMALL}$ & 62M  & 2,915           & 31,081       & 348.4s & 8.37 sentences/s  & 89.22 tokens/s \\
$\textrm{TrOCR}_{\rm BASE}$  & 334M & 2,915           & 31,959       & 633.7s & 4.60 sentences/s  & 50.43 tokens/s \\
$\textrm{TrOCR}_{\rm LARGE}$ & 558M & 2,915           & 31,966       & 666.8s & 4.37 sentences/s  & 47.94 tokens/s \\
\hline
\end{tabular}
\caption{Inference time on the IAM Handwriting dataset.}
\label{tab:infer}
\end{table*}

\begin{table*}[t]
  \small
  \centering
  \scalebox{1.02}{
  \setlength\tabcolsep{4pt}
  \begin{tabular*}{0.76\linewidth}{ c c c c c c c c c c }
    \toprule
    \multicolumn{1}{c}{} & \multicolumn{8}{c}{\bf Test datasets and \# of samples} \\
    \cmidrule{2-9}
    \multirow{2}{*}{\bf Model} & \bf IIIT5k & \bf SVT &  \multicolumn{2}{c}{\bf IC13} &  \multicolumn{2}{c}{\bf IC15} & \bf SVTP & \bf CUTE \\
     & 3,000 & 647 & 857 & 1,015 & 1,811 & 2,077 & 645 & 288 \\
    \midrule
        PlugNet \cite{mou2020plugnet} & 94.4 & 92.3 & -- & 95.0 & -- & 82.2 & 84.3 & 85.0 \\
        SRN \cite{9156632} & 94.8 & 91.5 & 95.5 & -- & 82.7 & -- & 85.1 & 87.8 \\
        RobustScanner \cite{yue2020robustscanner} & 95.4 & 89.3 & -- & 94.1 & -- & 79.2 & 82.9 & 92.4 \\
        TextScanner \cite{wan2020textscanner} & 95.7 & 92.7 & -- & 94.9 & -- & 83.5 & 84.8 & 91.6 \\
        AutoSTR \cite{zhang2020autostr} & 94.7 & 90.9 & -- & 94.2 & 81.8 & -- & 81.7 & -- \\
        RCEED \cite{cui_rceed} & 94.9 & 91.8 & -- & -- & -- & 82.2 & 83.6 & 91.7 \\
        PREN2D \cite{Yan_2021_CVPR} & 95.6 & 94.0 & 96.4 & -- & 83.0 & -- & 87.6 & 91.7 \\
        VisionLAN \cite{Wang_2021_ICCV_visionlan} & 95.8 & 91.7 & 95.7 & -- & 83.7 & -- & 86.0 & 88.5 \\
        Bhunia \cite{Bhunia_2021_ICCV_joint} & 95.2 & 92.2 & -- & 95.5 & -- & 84.0 & 85.7 & 89.7 \\
        CVAE-Feed.\textsuperscript{1} \cite{Bhunia_2021_ICCV_towards} & 95.2 & -- & -- & 95.7 & -- & \textbf{84.6} & 88.9 & 89.7 \\
        STN-CSTR \cite{https://doi.org/10.48550/arxiv.2102.10884} & 94.2 & 92.3 & 96.3 & 94.1 & 86.1 & 82.0 & 86.2 & -- \\
        ViTSTR-B \cite{atienza2021vision} & 88.4 & 87.7 & 93.2 & 92.4 & 78.5 & 72.6 & 81.8 & 81.3 \\
        CRNN \cite{shi2016end} & 84.3 & 78.9 & -- & 88.8 & -- & 61.5 & 64.8 & 61.3 \\
        TRBA \cite{Baek_2021_CVPR} & 92.1 & 88.9 & -- & 93.1 & -- & 74.7 & 79.5 & 78.2 \\
        ABINet \cite{Fang_2021_CVPR} & 96.2 & 93.5 & 97.4 & -- & 86.0 & -- & 89.3 & 89.2 \\
        Diaz \cite{diaz2021rethinking} & 96.8 & 94.6 & 96.0 & -- & 80.4 & -- & -- & -- \\
        PARSeq$_{A}$ \cite{bautista2022scene} & \textbf{97.0} & 93.6 & 97.0 & 96.2 & 86.5 & 82.9 & 88.9 & 92.2 \\
        MaskOCR (ViT-B) \cite{lyu2022maskocr} & 95.8 & 94.7 & 98.1 & - & 87.3 & - & 89.9 & 89.2 \\
        MaskOCR (ViT-L) \cite{lyu2022maskocr} & 96.5 & 94.1 & 97.8 & - & \textbf{88.7} & - & 90.2 & 92.7 \\
    \midrule
    $\textrm{TrOCR}_{\rm BASE}$ (Syn) & 90.1 & 91.0 & 97.3 & 96.3 & 81.1 & 75.0 & 90.7 & 86.8  \\
        $\textrm{TrOCR}_{\rm LARGE}$ (Syn) & 91.0 & 93.2 & 98.3 & 97.0 & 84.0 & 78.0 & 91.0 & 89.6  \\
        \midrule
        $\textrm{TrOCR}_{\rm BASE}$ (Syn+Benchmark) & 93.4 & 95.2 & 98.4 & \textbf{97.4} & 86.9 & 81.2 & 92.1 & 90.6  \\
       $\textrm{TrOCR}_{\rm LARGE}$ (Syn+Benchmark)& 94.1 & \textbf{96.1} & \textbf{98.4} & 97.3 & 88.1 & 84.1 & \textbf{93.0} & \textbf{95.1} \\
    \bottomrule
  \end{tabular*}
  }
  \caption{Word accuracy on the six benchmark datasets (36-char), where ``Syn'' indicates the model using synthetic data only and ``Syn+Benchmark'' indicates the model using synthetic data and benchmark datasets. }
  \label{tab:scene}
\end{table*}

\subsubsection{SROIE Task 2}

Table~\ref{tab:sroie} shows the results of the TrOCR models and the current SOTA methods on the leaderboard of the SROIE dataset.
To capture the visual information, all of these baselines leverage CNN-based networks as the feature extractors while the TrOCR models use the image Transformer to embed the information from the image patches. For language modeling, MSO Lab~\cite{sang2019improving} and CLOVA OCR~\cite{sang2019improving} use LSTM layers and H\&H Lab~\cite{shi2016end} use GRU layers while the TrOCR models use the Transformer decoder with a pure attention mechanism. According to the results, the TrOCR models outperform the existing SOTA models with pure Transformer structures. It is also confirmed that Transformer-based text recognition models get competitive performance compared to CNN-based networks in visual feature extraction and RNN-based networks in language modeling on this task without any complex pre/post-process steps.

\subsubsection{IAM Handwriting Database}
Table~\ref{tab:iam} shows the results of the TrOCR models and the existing methods on the IAM Handwriting Database. According to the results, the methods with CTC decoders show good performance on this task and the external LM will result in a significant reduction in CER. By comparing the methods \cite{bluche2017gated} with the TrOCR models, the $\textrm{TrOCR}_{\rm LARGE}$ achieves a better result, which indicates that the Transformer decoder is more competitive than the CTC decoder in text recognition and has enough ability for language modeling instead of relying on an external LM.
Most of the methods use sequence models in their encoders after the CNN-based backbone except the FCN encoders in \cite{wang2020decoupled}, which leads to a significant improvement on CER. Instead of relying on the features from the CNN-based backbone, the TrOCR models using the information from the image patches get similar and even better results, illustrating that the Transformer structures are competent to extract visual features well after pre-training. From the experiment results, the TrOCR models exceed all the methods which only use synthetic/IAM as the sole training set with pure Transformer structures and achieve a new state-of-the-art CER of 2.89. Without leveraging any extra human-labeled data, TrOCR even gets comparable results with the methods in \cite{diaz2021rethinking} using the additional internal human-labeled dataset. 

\subsubsection{Scene Text Datasets}
In Table~\ref{tab:scene}, we compare the $\textrm{TrOCR}_{\rm BASE}$ and $\textrm{TrOCR}_{\rm LARGE}$ models of fine-tuning with synthetic data only and fine-tuning with synthetic data and benchmark datasets (the training sets of IC13, IC15, IIIT5K, SVT) to the popular and recent SOTA methods. 
Compared to all, the TrOCR models establish five new SOTA results of eight experiments while getting comparable results on the rest. Our model underperforms on the IIIT5K dataset, and we find some scene text sample images contain symbols, but the ground truth does not. It is inconsistent with the behavior in our pre-training data (retaining symbols in ground truth), causing the model to tend still to process symbols. There are two kinds of mistakes: outputting symbols but truncating the output in advance to ensure that the number of wordpieces is consistent with the ground truth, or identifying symbols as similar characters.

\subsubsection{Inference Speed}
Table~\ref{tab:infer} shows the inference speed of different settings TrOCR models on the IAM Handwriting Database. We can conclude that there is no significant margin in inference speed between the base models and the large models. In contrast, the small model shows comparable results for printed and handwriting text recognition even though the number of parameters is an order of magnitude smaller and the inference speed is as twice as fast. The low number of parameters and high inference speed means fewer computational resources and user waiting time, making it more suitable for deployment in industrial applications.




\section{Related Work}

\subsection{Scene Text Recognition}
For text recognition, the most popular approaches are usually based on the CTC-based models. \cite{shi2016end} proposed the standard CRNN, an end-to-end architecture combined by CNN and RNN. The convolutional layers are used to extract the visual features and convert them to sequence by concatenating the columns, while the recurrent layers predict the per-frame labels. They use a CTC decoding strategy to remove the repeated symbols and all the blanks from the labels to achieve the final prediction. 
\cite{su2014accurate} used the Histogram of Oriented Gradient (HOG) features extracted from the image patches in the same column of the input image, instead of the features from the CNN network. A BiLSTM is then trained for labeling the sequential data with the CTC technique to find the best match.
\cite{gao2019reading} extracted the feature by the densely connected network incorporating the residual attention block and capture the contextual information and sequential dependency by the CNN network. They compute the probability distribution on the output of the CNN network instead of using an RNN network to model them. After that, CTC translates the probability distributions into the final label sequence. 

The Sequence-to-Sequence models \cite{zhang2020sahan, wang2019scene, sheng2019nrtr, bleeker2019bidirectional, lee2020recognizing, atienza2021vision} are gradually attracting more attention, especially after the advent of the Transformer architecture~\cite{vaswani2017attention}. 
SaHAN \cite{zhang2020sahan}, standing for the scale-aware hierarchical attention network, are proposed to address the character scale-variation issue. The authors use the FPN network and the CRNN models as the encoder as well as a hierarchical attention decoder to retain the multi-scale features. 
\cite{wang2019scene} extracted a sequence of visual features from the input images by the CNN with attention module and BiLSTM. The decoder is composed of the proposed Gated Cascade Attention Module (GCAM) and generates the target characters from the feature sequence extracted by the encoder.
For the Transformer models, \cite{sheng2019nrtr} first applied the Transformer to Scene Text Recognition. Since the input of the Transformer architecture is required to be a sequence, a CNN-based modality-transform block is employed to transform 2D input images to 1D sequences.
\cite{bleeker2019bidirectional} added a direction embedding to the input of the decoder for the bidirectional text decoding with a single decoder, while \cite{lee2020recognizing} utilized the two-dimensional dynamic positional embedding to keep the spatial structures of the intermediate feature maps for recognizing texts with arbitrary arrangements and large inter-character spacing. \cite{9156632} proposed semantic reasoning networks to replace RNN-like structures for more accurate text recognition.
\cite{atienza2021vision} only used the image Transformer without text Transformer for the text recognition in a non-autoregressive way.

The texts in natural images may appear in irregular shapes caused by perspective distortion. \cite{shi2016robust, baek2019wrong, litman2020scatter, shi2018aster, zhan2019esir} addressed this problem by processing the input images with an initial rectification step. For example, thin-plate spline transformation \cite{shi2016robust, baek2019wrong, litman2020scatter, shi2018aster} is applied to find a smooth spline interpolation between a set of fiducial points and normalize the text region to a predefined rectangle, while \cite{zhan2019esir} proposed an iterative rectification network to model the middle line of scene texts as well as the orientation and boundary of textlines. \cite{baek2019wrong, diaz2021rethinking} proposed universal architectures for comparing different recognition models. 

\subsection{Handwritten Text Recognition}
\cite{memon2020handwritten} gave a systematic literature review about the modern methods for handwriting recognition. Various attention mechanisms and positional encodings are compared in the \cite{michael2019evaluating} to address the alignment between the input and output sequence. The combination of RNN encoders (mostly LSTM) and CTC decoders \cite{bluche2017gated, graves2008offline, pham2014dropout} took a large part in the related works for a long time. Besides, \cite{graves2008offline, voigtlaender2016handwriting, puigcerver2017multidimensional} have also tried multidimensional LSTM encoders. Similar to the scene text recognition, the seq2seq methods and the scheme for attention decoding have been verified in \cite{michael2019evaluating, kang2020pay, chowdhury2018efficient, bluche2016joint}. \cite{ingle2019scalable} addressed the problems in building a large-scale system.

\section{Conclusion}

In this paper, we present TrOCR, an end-to-end Transformer-based OCR model for text recognition with pre-trained models. Distinct from existing approaches, TrOCR does not rely on the conventional CNN models for image understanding. Instead, it leverages an image Transformer model as the visual encoder and a text Transformer model as the textual decoder. Moreover, we use the wordpiece as the basic unit for the recognized output instead of the character-based methods, which saves the computational cost introduced by the additional language modeling. Experiment results show that TrOCR achieves state-of-the-art results on printed, handwritten and scene text recognition with just a simple encoder-decoder model, without any post-processing steps. 


\bibliography{aaai23}

\begin{thebibliography}{69}
\providecommand{\natexlab}[1]{#1}

\bibitem[{Atienza(2021)}]{atienza2021vision}
Atienza, R. 2021.
\newblock Vision Transformer for Fast and Efficient Scene Text Recognition.
\newblock \emph{arXiv preprint arXiv:2105.08582}.

\bibitem[{Baek et~al.(2019)Baek, Kim, Lee, Park, Han, Yun, Oh, and
  Lee}]{baek2019wrong}
Baek, J.; Kim, G.; Lee, J.; Park, S.; Han, D.; Yun, S.; Oh, S.~J.; and Lee, H.
  2019.
\newblock What is wrong with scene text recognition model comparisons? dataset
  and model analysis.
\newblock In \emph{Proceedings of the IEEE/CVF International Conference on
  Computer Vision}, 4715--4723.

\bibitem[{Baek, Matsui, and Aizawa(2021)}]{Baek_2021_CVPR}
Baek, J.; Matsui, Y.; and Aizawa, K. 2021.
\newblock What if We Only Use Real Datasets for Scene Text Recognition? Toward
  Scene Text Recognition With Fewer Labels.
\newblock In \emph{Proceedings of the IEEE/CVF Conference on Computer Vision
  and Pattern Recognition (CVPR)}, 3113--3122.

\bibitem[{Bao, Dong, and Wei(2021)}]{bao2021beit}
Bao, H.; Dong, L.; and Wei, F. 2021.
\newblock BEiT: BERT Pre-Training of Image Transformers.
\newblock arXiv:2106.08254.

\bibitem[{Bautista and Atienza(2022)}]{bautista2022scene}
Bautista, D.; and Atienza, R. 2022.
\newblock Scene Text Recognition with Permuted Autoregressive Sequence Models.
\newblock \emph{arXiv preprint arXiv:2207.06966}.

\bibitem[{Bhunia et~al.(2021{\natexlab{a}})Bhunia, Chowdhury, Sain, and
  Song}]{Bhunia_2021_ICCV_towards}
Bhunia, A.~K.; Chowdhury, P.~N.; Sain, A.; and Song, Y.-Z. 2021{\natexlab{a}}.
\newblock Towards the Unseen: Iterative Text Recognition by Distilling From
  Errors.
\newblock In \emph{Proceedings of the IEEE/CVF International Conference on
  Computer Vision (ICCV)}, 14950--14959.

\bibitem[{Bhunia et~al.(2021{\natexlab{b}})Bhunia, Sain, Kumar, Ghose,
  Chowdhury, and Song}]{Bhunia_2021_ICCV_joint}
Bhunia, A.~K.; Sain, A.; Kumar, A.; Ghose, S.; Chowdhury, P.~N.; and Song,
  Y.-Z. 2021{\natexlab{b}}.
\newblock Joint Visual Semantic Reasoning: Multi-Stage Decoder for Text
  Recognition.
\newblock In \emph{Proceedings of the IEEE/CVF International Conference on
  Computer Vision (ICCV)}, 14940--14949.

\bibitem[{Bleeker and de~Rijke(2019)}]{bleeker2019bidirectional}
Bleeker, M.; and de~Rijke, M. 2019.
\newblock Bidirectional scene text recognition with a single decoder.
\newblock \emph{arXiv preprint arXiv:1912.03656}.

\bibitem[{Bluche(2016)}]{bluche2016joint}
Bluche, T. 2016.
\newblock Joint line segmentation and transcription for end-to-end handwritten
  paragraph recognition.
\newblock \emph{Advances in Neural Information Processing Systems}, 29:
  838--846.

\bibitem[{Bluche and Messina(2017)}]{bluche2017gated}
Bluche, T.; and Messina, R. 2017.
\newblock Gated convolutional recurrent neural networks for multilingual
  handwriting recognition.
\newblock In \emph{2017 14th IAPR international conference on document analysis
  and recognition (ICDAR)}, volume~1, 646--651. IEEE.

\bibitem[{Cai, Sun, and
  Xiong(2021)}]{https://doi.org/10.48550/arxiv.2102.10884}
Cai, H.; Sun, J.; and Xiong, Y. 2021.
\newblock Revisiting Classification Perspective on Scene Text Recognition.

\bibitem[{Chowdhury and Vig(2018)}]{chowdhury2018efficient}
Chowdhury, A.; and Vig, L. 2018.
\newblock An efficient end-to-end neural model for handwritten text
  recognition.
\newblock \emph{arXiv preprint arXiv:1807.07965}.

\bibitem[{Cubuk et~al.(2020)Cubuk, Zoph, Shlens, and Le}]{cubuk2020randaugment}
Cubuk, E.~D.; Zoph, B.; Shlens, J.; and Le, Q.~V. 2020.
\newblock Randaugment: Practical automated data augmentation with a reduced
  search space.
\newblock In \emph{Proceedings of the IEEE/CVF conference on computer vision
  and pattern recognition workshops}, 702--703.

\bibitem[{Cui et~al.(2021)Cui, Wang, Zhang, and Wang}]{cui_rceed}
Cui, M.; Wang, W.; Zhang, J.; and Wang, L. 2021.
\newblock Representation and Correlation Enhanced Encoder-Decoder Framework for
  Scene Text Recognition.
\newblock In Llad{\'o}s, J.; Lopresti, D.; and Uchida, S., eds., \emph{Document
  Analysis and Recognition -- ICDAR 2021}, 156--170. Cham: Springer
  International Publishing.
\newblock ISBN 978-3-030-86337-1.

\bibitem[{Deng et~al.(2009)Deng, Dong, Socher, Li, Li, and
  Fei-Fei}]{deng2009imagenet}
Deng, J.; Dong, W.; Socher, R.; Li, L.-J.; Li, K.; and Fei-Fei, L. 2009.
\newblock Imagenet: A large-scale hierarchical image database.
\newblock In \emph{2009 IEEE conference on computer vision and pattern
  recognition}, 248--255. Ieee.

\bibitem[{Devlin et~al.(2019)Devlin, Chang, Lee, and
  Toutanova}]{devlin2019bert}
Devlin, J.; Chang, M.-W.; Lee, K.; and Toutanova, K. 2019.
\newblock BERT: Pre-training of Deep Bidirectional Transformers for Language
  Understanding.
\newblock arXiv:1810.04805.

\bibitem[{Diaz et~al.(2021)Diaz, Qin, Ingle, Fujii, and
  Bissacco}]{diaz2021rethinking}
Diaz, D.~H.; Qin, S.; Ingle, R.; Fujii, Y.; and Bissacco, A. 2021.
\newblock Rethinking Text Line Recognition Models.
\newblock arXiv:2104.07787.

\bibitem[{Dong et~al.(2019)Dong, Yang, Wang, Wei, Liu, Wang, Gao, Zhou, and
  Hon}]{dong2019unified}
Dong, L.; Yang, N.; Wang, W.; Wei, F.; Liu, X.; Wang, Y.; Gao, J.; Zhou, M.;
  and Hon, H.-W. 2019.
\newblock Unified Language Model Pre-training for Natural Language
  Understanding and Generation.
\newblock arXiv:1905.03197.

\bibitem[{Dosovitskiy et~al.(2021)Dosovitskiy, Beyer, Kolesnikov, Weissenborn,
  Zhai, Unterthiner, Dehghani, Minderer, Heigold, Gelly, Uszkoreit, and
  Houlsby}]{dosovitskiy2020vit}
Dosovitskiy, A.; Beyer, L.; Kolesnikov, A.; Weissenborn, D.; Zhai, X.;
  Unterthiner, T.; Dehghani, M.; Minderer, M.; Heigold, G.; Gelly, S.;
  Uszkoreit, J.; and Houlsby, N. 2021.
\newblock An Image is Worth 16x16 Words: Transformers for Image Recognition at
  Scale.
\newblock \emph{ICLR}.

\bibitem[{Fang et~al.(2021)Fang, Xie, Wang, Mao, and Zhang}]{Fang_2021_CVPR}
Fang, S.; Xie, H.; Wang, Y.; Mao, Z.; and Zhang, Y. 2021.
\newblock Read Like Humans: Autonomous, Bidirectional and Iterative Language
  Modeling for Scene Text Recognition.
\newblock In \emph{Proceedings of the IEEE/CVF Conference on Computer Vision
  and Pattern Recognition (CVPR)}, 7098--7107.

\bibitem[{Gao et~al.(2019)Gao, Chen, Wang, Tang, and Lu}]{gao2019reading}
Gao, Y.; Chen, Y.; Wang, J.; Tang, M.; and Lu, H. 2019.
\newblock Reading scene text with fully convolutional sequence modeling.
\newblock \emph{Neurocomputing}, 339: 161--170.

\bibitem[{Graves et~al.(2006)Graves, Fern{\'a}ndez, Gomez, and
  Schmidhuber}]{graves2006connectionist}
Graves, A.; Fern{\'a}ndez, S.; Gomez, F.; and Schmidhuber, J. 2006.
\newblock Connectionist temporal classification: labelling unsegmented sequence
  data with recurrent neural networks.
\newblock In \emph{Proceedings of the 23rd international conference on Machine
  learning}, 369--376.

\bibitem[{Graves and Schmidhuber(2008)}]{graves2008offline}
Graves, A.; and Schmidhuber, J. 2008.
\newblock Offline handwriting recognition with multidimensional recurrent
  neural networks.
\newblock \emph{Advances in neural information processing systems}, 21:
  545--552.

\bibitem[{Gupta, Vedaldi, and Zisserman(2016)}]{gupta2016synthetic}
Gupta, A.; Vedaldi, A.; and Zisserman, A. 2016.
\newblock Synthetic data for text localisation in natural images.
\newblock In \emph{Proceedings of the IEEE conference on computer vision and
  pattern recognition}, 2315--2324.

\bibitem[{Ingle et~al.(2019)Ingle, Fujii, Deselaers, Baccash, and
  Popat}]{ingle2019scalable}
Ingle, R.~R.; Fujii, Y.; Deselaers, T.; Baccash, J.; and Popat, A.~C. 2019.
\newblock A scalable handwritten text recognition system.
\newblock In \emph{2019 International Conference on Document Analysis and
  Recognition (ICDAR)}, 17--24. IEEE.

\bibitem[{Jaderberg et~al.(2014)Jaderberg, Simonyan, Vedaldi, and
  Zisserman}]{synth90ka}
Jaderberg, M.; Simonyan, K.; Vedaldi, A.; and Zisserman, A. 2014.
\newblock Synthetic Data and Artificial Neural Networks for Natural Scene Text
  Recognition.
\newblock In \emph{Workshop on Deep Learning, NIPS}.

\bibitem[{Kang et~al.(2020)Kang, Riba, Rusi{\~n}ol, Forn{\'e}s, and
  Villegas}]{kang2020pay}
Kang, L.; Riba, P.; Rusi{\~n}ol, M.; Forn{\'e}s, A.; and Villegas, M. 2020.
\newblock Pay attention to what you read: Non-recurrent handwritten text-line
  recognition.
\newblock \emph{arXiv preprint arXiv:2005.13044}.

\bibitem[{Karatzas et~al.(2015)Karatzas, Gomez-Bigorda, Nicolaou, Ghosh,
  Bagdanov, Iwamura, Matas, Neumann, Chandrasekhar, Lu
  et~al.}]{karatzas2015icdar}
Karatzas, D.; Gomez-Bigorda, L.; Nicolaou, A.; Ghosh, S.; Bagdanov, A.;
  Iwamura, M.; Matas, J.; Neumann, L.; Chandrasekhar, V.~R.; Lu, S.; et~al.
  2015.
\newblock ICDAR 2015 competition on robust reading.
\newblock In \emph{ICDAR}.

\bibitem[{Karatzas et~al.(2013)Karatzas, Shafait, Uchida, Iwamura, i~Bigorda,
  Mestre, Mas, Mota, Almazan, and De~Las~Heras}]{karatzas2013icdar}
Karatzas, D.; Shafait, F.; Uchida, S.; Iwamura, M.; i~Bigorda, L.~G.; Mestre,
  S.~R.; Mas, J.; Mota, D.~F.; Almazan, J.~A.; and De~Las~Heras, L.~P. 2013.
\newblock ICDAR 2013 robust reading competition.
\newblock In \emph{ICDAR}.

\bibitem[{Krishnan and Jawahar(2016)}]{krishnan2016generating}
Krishnan, P.; and Jawahar, C.~V. 2016.
\newblock Generating Synthetic Data for Text Recognition.
\newblock arXiv:1608.04224.

\bibitem[{Kudo and Richardson(2018)}]{kudo2018sentencepiece}
Kudo, T.; and Richardson, J. 2018.
\newblock Sentencepiece: A simple and language independent subword tokenizer
  and detokenizer for neural text processing.
\newblock \emph{arXiv preprint arXiv:1808.06226}.

\bibitem[{Lee et~al.(2020)Lee, Park, Baek, Oh, Kim, and
  Lee}]{lee2020recognizing}
Lee, J.; Park, S.; Baek, J.; Oh, S.~J.; Kim, S.; and Lee, H. 2020.
\newblock On recognizing texts of arbitrary shapes with 2D self-attention.
\newblock In \emph{Proceedings of the IEEE/CVF Conference on Computer Vision
  and Pattern Recognition Workshops}, 546--547.

\bibitem[{Liao et~al.(2019)Liao, Wan, Yao, Chen, and Bai}]{liao2019realtime}
Liao, M.; Wan, Z.; Yao, C.; Chen, K.; and Bai, X. 2019.
\newblock Real-time Scene Text Detection with Differentiable Binarization.
\newblock arXiv:1911.08947.

\bibitem[{Litman et~al.(2020)Litman, Anschel, Tsiper, Litman, Mazor, and
  Manmatha}]{litman2020scatter}
Litman, R.; Anschel, O.; Tsiper, S.; Litman, R.; Mazor, S.; and Manmatha, R.
  2020.
\newblock Scatter: selective context attentional scene text recognizer.
\newblock In \emph{Proceedings of the IEEE/CVF Conference on Computer Vision
  and Pattern Recognition}, 11962--11972.

\bibitem[{Liu et~al.(2019)Liu, Ott, Goyal, Du, Joshi, Chen, Levy, Lewis,
  Zettlemoyer, and Stoyanov}]{liu2019roberta}
Liu, Y.; Ott, M.; Goyal, N.; Du, J.; Joshi, M.; Chen, D.; Levy, O.; Lewis, M.;
  Zettlemoyer, L.; and Stoyanov, V. 2019.
\newblock RoBERTa: A Robustly Optimized BERT Pretraining Approach.
\newblock arXiv:1907.11692.

\bibitem[{Lyu et~al.(2022)Lyu, Zhang, Liu, Qiao, Xu, Wu, Yao, Han, Ding, and
  Wang}]{lyu2022maskocr}
Lyu, P.; Zhang, C.; Liu, S.; Qiao, M.; Xu, Y.; Wu, L.; Yao, K.; Han, J.; Ding,
  E.; and Wang, J. 2022.
\newblock MaskOCR: Text Recognition with Masked Encoder-Decoder Pretraining.
\newblock \emph{arXiv preprint arXiv:2206.00311}.

\bibitem[{Memon et~al.(2020)Memon, Sami, Khan, and
  Uddin}]{memon2020handwritten}
Memon, J.; Sami, M.; Khan, R.~A.; and Uddin, M. 2020.
\newblock Handwritten optical character recognition (OCR): A comprehensive
  systematic literature review (SLR).
\newblock \emph{IEEE Access}, 8: 142642--142668.

\bibitem[{Michael et~al.(2019)Michael, Labahn, Gr{\"u}ning, and
  Z{\"o}llner}]{michael2019evaluating}
Michael, J.; Labahn, R.; Gr{\"u}ning, T.; and Z{\"o}llner, J. 2019.
\newblock Evaluating sequence-to-sequence models for handwritten text
  recognition.
\newblock In \emph{2019 International Conference on Document Analysis and
  Recognition (ICDAR)}, 1286--1293. IEEE.

\bibitem[{Mishra, Alahari, and Jawahar(2012)}]{mishra2012top}
Mishra, A.; Alahari, K.; and Jawahar, C. 2012.
\newblock Top-down and bottom-up cues for scene text recognition.
\newblock In \emph{CVPR}.

\bibitem[{Mou et~al.(2020)Mou, Tan, Yang, Chen, Liu, Yan, and
  Huang}]{mou2020plugnet}
Mou, Y.; Tan, L.; Yang, H.; Chen, J.; Liu, L.; Yan, R.; and Huang, Y. 2020.
\newblock Plugnet: Degradation aware scene text recognition supervised by a
  pluggable super-resolution unit.
\newblock In \emph{Computer Vision--ECCV 2020: 16th European Conference,
  Glasgow, UK, August 23--28, 2020, Proceedings, Part XV 16}, 158--174.
  Springer.

\bibitem[{Ott et~al.(2019)Ott, Edunov, Baevski, Fan, Gross, Ng, Grangier, and
  Auli}]{ott2019fairseq}
Ott, M.; Edunov, S.; Baevski, A.; Fan, A.; Gross, S.; Ng, N.; Grangier, D.; and
  Auli, M. 2019.
\newblock fairseq: A Fast, Extensible Toolkit for Sequence Modeling.
\newblock In \emph{Proceedings of NAACL-HLT 2019: Demonstrations}.

\bibitem[{Pham et~al.(2014)Pham, Bluche, Kermorvant, and
  Louradour}]{pham2014dropout}
Pham, V.; Bluche, T.; Kermorvant, C.; and Louradour, J. 2014.
\newblock Dropout improves recurrent neural networks for handwriting
  recognition.
\newblock In \emph{2014 14th international conference on frontiers in
  handwriting recognition}, 285--290. IEEE.

\bibitem[{Phan et~al.(2013)Phan, Shivakumara, Tian, and
  Tan}]{phan2013recognizing}
Phan, T.~Q.; Shivakumara, P.; Tian, S.; and Tan, C.~L. 2013.
\newblock Recognizing text with perspective distortion in natural scenes.
\newblock In \emph{Proceedings of the IEEE International Conference on Computer
  Vision}, 569--576.

\bibitem[{Puigcerver(2017)}]{puigcerver2017multidimensional}
Puigcerver, J. 2017.
\newblock Are multidimensional recurrent layers really necessary for
  handwritten text recognition?
\newblock In \emph{2017 14th IAPR International Conference on Document Analysis
  and Recognition (ICDAR)}, volume~1, 67--72. IEEE.

\bibitem[{Ramesh et~al.(2021)Ramesh, Pavlov, Goh, Gray, Voss, Radford, Chen,
  and Sutskever}]{ramesh2021zero}
Ramesh, A.; Pavlov, M.; Goh, G.; Gray, S.; Voss, C.; Radford, A.; Chen, M.; and
  Sutskever, I. 2021.
\newblock Zero-shot text-to-image generation.
\newblock \emph{arXiv preprint arXiv:2102.12092}.

\bibitem[{Risnumawan et~al.(2014)Risnumawan, Shivakumara, Chan, and
  Tan}]{risnumawan2014robust}
Risnumawan, A.; Shivakumara, P.; Chan, C.~S.; and Tan, C.~L. 2014.
\newblock A robust arbitrary text detection system for natural scene images.
\newblock \emph{Expert Systems with Applications}.

\bibitem[{Sang and Cuong(2019)}]{sang2019improving}
Sang, D.~V.; and Cuong, L. T.~B. 2019.
\newblock Improving CRNN with EfficientNet-like feature extractor and
  multi-head attention for text recognition.
\newblock In \emph{Proceedings of the Tenth International Symposium on
  Information and Communication Technology}, 285--290.

\bibitem[{Sennrich, Haddow, and Birch(2015)}]{sennrich2015neural}
Sennrich, R.; Haddow, B.; and Birch, A. 2015.
\newblock Neural machine translation of rare words with subword units.
\newblock \emph{arXiv preprint arXiv:1508.07909}.

\bibitem[{Sheng, Chen, and Xu(2019)}]{sheng2019nrtr}
Sheng, F.; Chen, Z.; and Xu, B. 2019.
\newblock NRTR: A no-recurrence sequence-to-sequence model for scene text
  recognition.
\newblock In \emph{2019 International Conference on Document Analysis and
  Recognition (ICDAR)}, 781--786. IEEE.

\bibitem[{Shi, Bai, and Yao(2016)}]{shi2016end}
Shi, B.; Bai, X.; and Yao, C. 2016.
\newblock An end-to-end trainable neural network for image-based sequence
  recognition and its application to scene text recognition.
\newblock \emph{IEEE transactions on pattern analysis and machine
  intelligence}, 39(11): 2298--2304.

\bibitem[{Shi et~al.(2016)Shi, Wang, Lyu, Yao, and Bai}]{shi2016robust}
Shi, B.; Wang, X.; Lyu, P.; Yao, C.; and Bai, X. 2016.
\newblock Robust scene text recognition with automatic rectification.
\newblock In \emph{Proceedings of the IEEE conference on computer vision and
  pattern recognition}, 4168--4176.

\bibitem[{Shi et~al.(2018)Shi, Yang, Wang, Lyu, Yao, and Bai}]{shi2018aster}
Shi, B.; Yang, M.; Wang, X.; Lyu, P.; Yao, C.; and Bai, X. 2018.
\newblock Aster: An attentional scene text recognizer with flexible
  rectification.
\newblock \emph{IEEE transactions on pattern analysis and machine
  intelligence}, 41(9): 2035--2048.

\bibitem[{Su and Lu(2014)}]{su2014accurate}
Su, B.; and Lu, S. 2014.
\newblock Accurate scene text recognition based on recurrent neural network.
\newblock In \emph{Asian Conference on Computer Vision}, 35--48. Springer.

\bibitem[{Touvron et~al.(2021)Touvron, Cord, Douze, Massa, Sablayrolles, and
  J{\'e}gou}]{touvron2020deit}
Touvron, H.; Cord, M.; Douze, M.; Massa, F.; Sablayrolles, A.; and J{\'e}gou,
  H. 2021.
\newblock Training data-efficient image transformers \& distillation through
  attention.
\newblock In \emph{International Conference on Machine Learning}, 10347--10357.
  PMLR.

\bibitem[{Vaswani et~al.(2017)Vaswani, Shazeer, Parmar, Uszkoreit, Jones,
  Gomez, Kaiser, and Polosukhin}]{vaswani2017attention}
Vaswani, A.; Shazeer, N.; Parmar, N.; Uszkoreit, J.; Jones, L.; Gomez, A.~N.;
  Kaiser, {\L}.; and Polosukhin, I. 2017.
\newblock Attention is all you need.
\newblock In \emph{Advances in neural information processing systems},
  5998--6008.

\bibitem[{Voigtlaender, Doetsch, and Ney(2016)}]{voigtlaender2016handwriting}
Voigtlaender, P.; Doetsch, P.; and Ney, H. 2016.
\newblock Handwriting recognition with large multidimensional long short-term
  memory recurrent neural networks.
\newblock In \emph{2016 15th International Conference on Frontiers in
  Handwriting Recognition (ICFHR)}, 228--233. IEEE.

\bibitem[{Wan et~al.(2020)Wan, He, Chen, Bai, and Yao}]{wan2020textscanner}
Wan, Z.; He, M.; Chen, H.; Bai, X.; and Yao, C. 2020.
\newblock Textscanner: Reading characters in order for robust scene text
  recognition.
\newblock In \emph{Proceedings of the AAAI Conference on Artificial
  Intelligence}, volume~34, 12120--12127.

\bibitem[{Wang, Babenko, and Belongie(2011)}]{wang2011end}
Wang, K.; Babenko, B.; and Belongie, S. 2011.
\newblock End-to-end scene text recognition.
\newblock In \emph{2011 International conference on computer vision},
  1457--1464. IEEE.

\bibitem[{Wang et~al.(2019)Wang, Wang, Qin, Zhao, and Tang}]{wang2019scene}
Wang, S.; Wang, Y.; Qin, X.; Zhao, Q.; and Tang, Z. 2019.
\newblock Scene text recognition via gated cascade attention.
\newblock In \emph{2019 IEEE International Conference on Multimedia and Expo
  (ICME)}, 1018--1023. IEEE.

\bibitem[{Wang et~al.(2020{\natexlab{a}})Wang, Zhu, Jin, Luo, Chen, Wu, Wang,
  and Cai}]{wang2020decoupled}
Wang, T.; Zhu, Y.; Jin, L.; Luo, C.; Chen, X.; Wu, Y.; Wang, Q.; and Cai, M.
  2020{\natexlab{a}}.
\newblock Decoupled Attention Network for Text Recognition.
\newblock In \emph{Proceedings of the AAAI Conference on Artificial
  Intelligence}.

\bibitem[{Wang et~al.(2020{\natexlab{b}})Wang, Wei, Dong, Bao, Yang, and
  Zhou}]{wang2020minilm}
Wang, W.; Wei, F.; Dong, L.; Bao, H.; Yang, N.; and Zhou, M.
  2020{\natexlab{b}}.
\newblock Minilm: Deep self-attention distillation for task-agnostic
  compression of pre-trained transformers.
\newblock \emph{arXiv preprint arXiv:2002.10957}.

\bibitem[{Wang et~al.(2021)Wang, Xie, Fang, Wang, Zhu, and
  Zhang}]{Wang_2021_ICCV_visionlan}
Wang, Y.; Xie, H.; Fang, S.; Wang, J.; Zhu, S.; and Zhang, Y. 2021.
\newblock From Two to One: A New Scene Text Recognizer With Visual Language
  Modeling Network.
\newblock In \emph{Proceedings of the IEEE/CVF International Conference on
  Computer Vision (ICCV)}, 14194--14203.

\bibitem[{Wightman(2019)}]{rw2019timm}
Wightman, R. 2019.
\newblock PyTorch Image Models.
\newblock \url{https://github.com/rwightman/pytorch-image-models}.

\bibitem[{Yan et~al.(2021)Yan, Peng, Xiao, and Yao}]{Yan_2021_CVPR}
Yan, R.; Peng, L.; Xiao, S.; and Yao, G. 2021.
\newblock Primitive Representation Learning for Scene Text Recognition.
\newblock In \emph{Proceedings of the IEEE/CVF Conference on Computer Vision
  and Pattern Recognition (CVPR)}, 284--293.

\bibitem[{Yu et~al.(2020)Yu, Li, Zhang, Liu, Han, Liu, and Ding}]{9156632}
Yu, D.; Li, X.; Zhang, C.; Liu, T.; Han, J.; Liu, J.; and Ding, E. 2020.
\newblock Towards Accurate Scene Text Recognition With Semantic Reasoning
  Networks.
\newblock In \emph{2020 IEEE/CVF Conference on Computer Vision and Pattern
  Recognition (CVPR)}, 12110--12119.

\bibitem[{Yue et~al.(2020)Yue, Kuang, Lin, Sun, and
  Zhang}]{yue2020robustscanner}
Yue, X.; Kuang, Z.; Lin, C.; Sun, H.; and Zhang, W. 2020.
\newblock Robustscanner: Dynamically enhancing positional clues for robust text
  recognition.
\newblock In \emph{European Conference on Computer Vision}, 135--151. Springer.

\bibitem[{Zhan and Lu(2019)}]{zhan2019esir}
Zhan, F.; and Lu, S. 2019.
\newblock Esir: End-to-end scene text recognition via iterative image
  rectification.
\newblock In \emph{Proceedings of the IEEE/CVF Conference on Computer Vision
  and Pattern Recognition}, 2059--2068.

\bibitem[{Zhang et~al.(2020{\natexlab{a}})Zhang, Yao, Yang, Xu, and
  Bai}]{zhang2020autostr}
Zhang, H.; Yao, Q.; Yang, M.; Xu, Y.; and Bai, X. 2020{\natexlab{a}}.
\newblock AutoSTR: Efficient backbone search for scene text recognition.
\newblock In \emph{Computer Vision--ECCV 2020: 16th European Conference,
  Glasgow, UK, August 23--28, 2020, Proceedings, Part XXIV 16}, 751--767.
  Springer.

\bibitem[{Zhang et~al.(2020{\natexlab{b}})Zhang, Luo, Jin, Wang, Li, and
  Zhou}]{zhang2020sahan}
Zhang, J.; Luo, C.; Jin, L.; Wang, T.; Li, Z.; and Zhou, W. 2020{\natexlab{b}}.
\newblock SaHAN: Scale-aware hierarchical attention network for scene text
  recognition.
\newblock \emph{Pattern Recognition Letters}, 136: 205--211.

\end{thebibliography}

\end{document}